\documentclass[pdflatex,sn-mathphys-num]{sn-jnl}

\usepackage{graphicx}%
\usepackage{multirow}%
\usepackage{amsmath,amssymb,amsfonts}%
\usepackage{amsthm}%
\usepackage{mathrsfs}%
\usepackage[title]{appendix}%
\usepackage{xcolor}%
\usepackage{textcomp}%
\usepackage{manyfoot}%
\usepackage{booktabs}%
\usepackage{algorithm}%
\usepackage{algorithmicx}%
\usepackage{algpseudocode}%
\usepackage{listings}%
\usepackage{graphicx}
\usepackage{multirow}
\usepackage{amsmath,amssymb,amsfonts}
\usepackage{amsthm}
\usepackage{mathrsfs}
\usepackage[title]{appendix}
\usepackage{xcolor}
\usepackage{textcomp}
\usepackage{manyfoot}
\usepackage{booktabs}
\usepackage{algorithm}
\usepackage{algorithmicx}
\usepackage{algpseudocode}
\usepackage{listings}
\usepackage{todonotes}

\theoremstyle{thmstyleone}%
\theoremstyle{thmstyletwo}%

\theoremstyle{thmstylethree}%

\begin{document}

\title[Kirchhoff-Inspired Neural Networks]{Circuit-Inspired High-Order Neural Networks with Unified Neural Dynamics Modeling for PDE Solving and Visual Perception }


\author[1]{\fnm{Tongfei} \sur{Chen}}
\equalcont{These authors contributed equally to this work.}

\author[1]{\fnm{Jingying} \sur{Yang}}
\equalcont{These authors contributed equally to this work.}

\author*[2]{\fnm{Linlin} \sur{Yang}}\email{lyang@cuc.edu.cn}

\author*[1]{\fnm{Juan} \sur{Zhang}}\email{zhang\_juan@buaa.edu.cn}

\author[3]{\fnm{Jinhu} \sur{L{\"u}}}

\author[4]{\fnm{David} \sur{Doermann}}

\author[5]{\fnm{Chunyu} \sur{Xie}}

\author[1]{\fnm{Long} \sur{He}}

\author[1]{\fnm{Tian} \sur{Wang}}

\author[6]{\fnm{Guodong} \sur{Guo}}

\author*[1]{\fnm{Baochang} \sur{Zhang}}\email{bczhang@buaa.edu.cn}

\affil[1]{\orgdiv{School of Artificial Intelligence (Institute of Artificial Intelligence)}, \orgname{Beihang University}, \orgaddress{\city{Beijing}, \country{China}}}

\affil[2]{\orgname{Communication University of China}, \orgaddress{\city{Beijing}, \country{China}}}

\affil[3]{\orgdiv{School of Automation Science and Electrical Engineering}, \orgname{Beihang University}, \orgaddress{\city{Beijing}, \country{China}}}

\affil[4]{\orgdiv{Department of Computer Science and Engineering}, \orgname{University at Buffalo}, \orgaddress{\city{Buffalo}, \state{New York}, \country{USA}}}

\affil[5]{\orgname{360 AI Research, Qihoo 360}, \orgaddress{\city{Beijing}, \country{China}}}

\affil[6]{\orgname{Eastern Institute of Technology, Ningbo}, \orgaddress{\city{Ningbo}, \country{China}}}

\abstract{
Deep networks often rely on architectural heuristics to shape representation evolution, limiting their ability to model data governed by intrinsic dynamics. We present the Circuit-inspired High-Order Neural Network (CHONN), a modular framework that treats representation evolution as a latent potential process and increases its effective order through Kirchhoff-inspired cascade composition. A single Kirchhoff Neural Cell implements a stable first-order update, while serially composed cells form higher-order dynamical operators within one block. This construction is interpretable, numerically stable and compatible with common neural backbones. Theoretical analysis shows that cascaded cells induce end-to-end high-order operators, and controlled experiments demonstrate that intra-block high-order construction differs from generic depth stacking, especially on derivative-sensitive measures. Across steady-state operator learning, long-horizon physical forecasting and ImageNet-1K recognition, CHONN improves structural fidelity, rollout stability and visual representation learning. These results identify high-order circuit composition as a general principle for neural dynamics modeling.
}

\keywords{circuit-inspired neural networks, PDE operator learning, visual representation learning}



\maketitle

Modern deep networks are highly effective in learning useful representations~\cite{bengio2013representation,krizhevsky2012imagenet,he2016resnet,vaswani2017attention,dosovitskiy2021image}.  However, representation evolution,  governed by intrinsic system dynamics during  embedding learning, is often overlooked when modeling  such a dynamical process~\cite{chen2018neural}.
In the biological sensory pathway, information should be carried by continuous membrane-potential transformations across connected neural populations~\cite{gerstner2014neuronal,dayan2001theoretical,debanne2013analog,averbeck2006neural,vyas2020computation}, illustrating how representation evolution can be embedded within intrinsic system dynamics. From this perspective, neural computation depends not only on signal strength and  interaction structure, but also on intrinsic representation evolution.

\begin{figure}[t]
\centering
\includegraphics[width=1.0\textwidth]{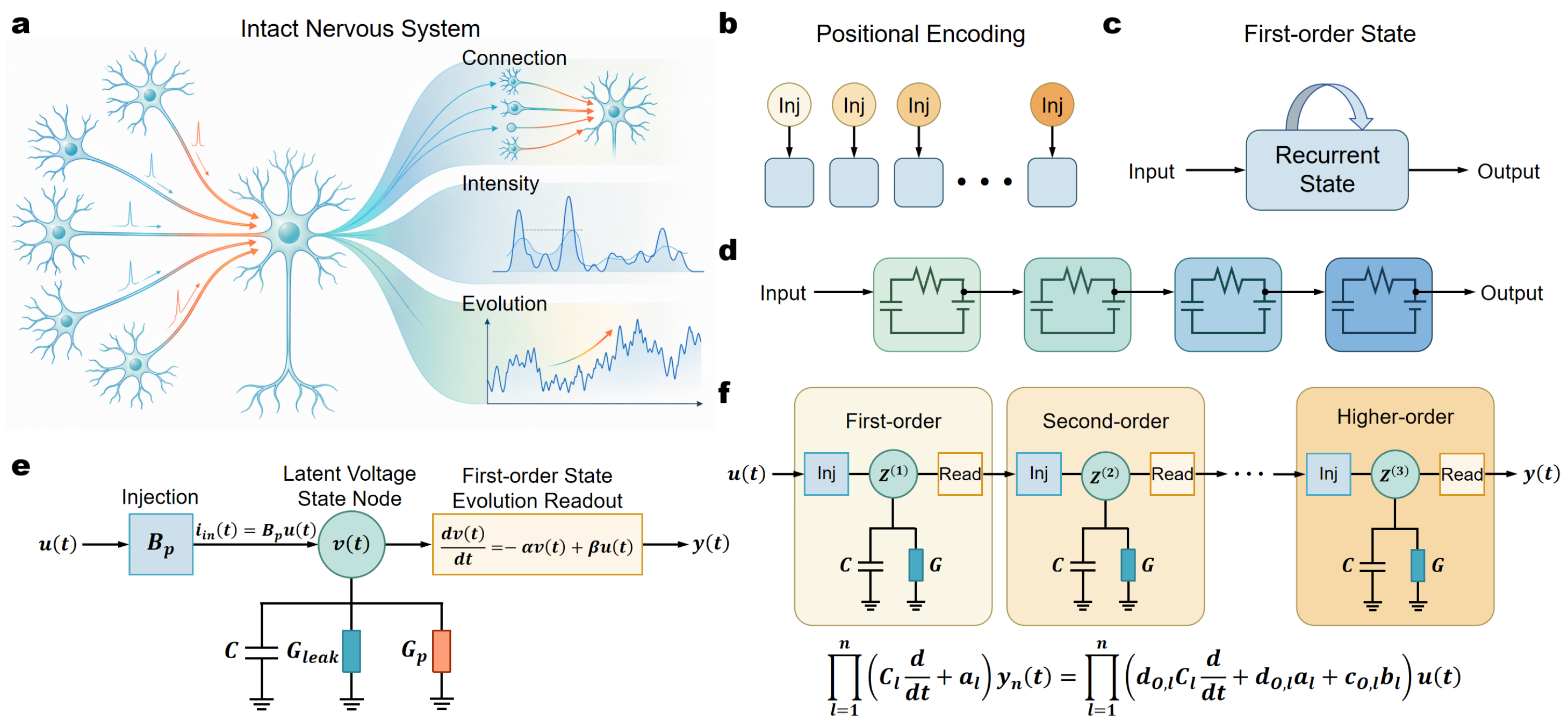}
\caption{
\textbf{CHONN builds high-order representation evolution through Kirchhoff-inspired cascade composition.}
\textbf{a,} Neural-population analogy motivating the separation of connection, intensity and evolution in signal propagation.
\textbf{b,} Positional encodings introduce variation across positions through externally supplied index-dependent cues.
\textbf{c,} Recurrent state models introduce an internal state, but a single transition corresponds to a first-order update.
\textbf{d,} CHONN composes circuit-inspired latent updates in series, allowing representation dynamics to accumulate across stages.
\textbf{e,} A Kirchhoff Neural Cell (KNC) represents the hidden feature as a latent potential driven by input injection, leakage and coupling, followed by output readout.
\textbf{f,} Cascading KNCs yields second- and higher-order evolution factors, making the order of latent representation dynamics explicit within one block.
}\label{fig1:intro}
\end{figure}

By contrast, in many contemporary feedforward and attention-based network architectures, changes in representation across positions or steps are typically introduced through architectural heuristics, such as positional encodings, attention masks, or gating mechanisms, rather than formulated as an internal representational evolution process~\cite{vaswani2017attention,shazeer2020glu,santos2023development}. 
{Although such mechanisms are often effective in practice, they do not make evolution itself an explicit carrier of representation~\cite{chen2018neural,hasani2022closed}. Recurrent and state-space models introduce internal states, but a single state transition usually provides only the first-order representation~\cite{gu2022s4,smith2022diagonal,gu2023mamba,liu2024vmamba,tiezzi2025back}}. As a consequence, existing methods are rarely designed to jointly capture input strength, interaction structure and high-order evolution within one unified dynamical framework. Order refers to the complexity level of the evolutionary process within a dynamical system, specifically, the order of the partial differential equation (PDE) describing the system. This limitation may be especially {consequential} for data governed by continuous physical dynamics, including PDEs. {{As shown in Fig.~\ref{fig1:intro}a-c, neural computation should jointly account for signal strength, interaction structure, and intrinsic representation evolution, whereas existing architectures either introduce evolution through architectural heuristics or restrict it to a first-order state transition.}}

We propose the Circuit-inspired High-Order Neural Network (CHONN), which represents evolution as an intrinsic latent potential process and elevates its order through cascade composition (Fig.~\ref{fig1:intro}d). Rather than appending positional or transition heuristics to otherwise near-instantaneous mappings, CHONN builds {zeroth-, first-, and higher-order} dynamics directly into the representation through  circuit-inspired latent updates. {Specifically, inspired by Kirchhoff's circuit laws, we model representation evolution using a Kirchhoff Neural Cell (KNC) for the first-order  representation under Resistor-Capacitor (RC) dynamics (Fig.~\ref{fig1:intro}e), and  a cascade of KNCs for higher-order representations (Fig.~\ref{fig1:intro}f).} The resulting framework is interpretable, numerically stable, and readily deployable across diverse neural backbones.

We support this view with both mechanism-level and task-level evidence. In order-defined task-level analyses, we show that the intra-block high-order mechanism
cannot degenerate to generic depth expansion: its advantages are consistent 
with increasing order and are most salient
on derivative-sensitive measures and scaling up. Across operator learning, long-horizon physical forecasting,  and large-scale visual recognition, CHONN yields consistent gains. At the same time, ablations show that these gains arise specifically from sequential high-order composition rather than from increased depth alone.

\section*{Results}

\subsection*{Cascaded Kirchhoff Neural Cells realize high-order circuit-inspired evolution}
\label{subsec:results_highorder_knc}

CHONN models representation evolution as a circuit-inspired latent-potential process rather than as externally supplied positional information~\cite{morita2024positional,zhao2024length}. At the single-cell level, a Kirchhoff Neural Cell (KNC) treats the hidden representation as a latent potential \(v(t)\) governed by RC dynamics~\cite{chua1987linear,nilsson2022electric}:
\begin{equation}
C\frac{d v(t)}{dt}=-(G_{\mathrm{leak}}+G_p)v(t)+B_pu(t).
\label{eq:results_rc_ode}
\end{equation}
Here, \(C\) denotes the effective capacitance, \(G_{\mathrm{leak}}\) and \(G_p\) denote leakage and coupling conductances, and \(B_pu(t)\) represents input-driven injection. With \(\alpha=(G_{\mathrm{leak}}+G_p)/C\) and \(\beta=B_p/C\), this becomes the first-order latent evolution
\begin{equation}
\dot v(t)=-\alpha v(t)+\beta u(t),
\end{equation}
where relaxation and input injection jointly determine the state update.

For neural implementation, this continuous-time dynamics is discretized under zero-order hold. Let \(v_k=v(t_k)\), \(u_k=u(t_k)\), and \(\Delta t_k=t_k-t_{k-1}\). Assuming the input is held constant within each interval gives the closed-form update
\begin{equation}
v_k=\bar\alpha_k v_{k-1}+\bar\beta_k u_k,
\quad
\bar\alpha_k=e^{-\alpha\Delta t_k},
\quad
\bar\beta_k=\frac{\beta}{\alpha}\left(1-e^{-\alpha\Delta t_k}\right).
\label{eq:results_zoh_exact}
\end{equation}
This form provides a stable recurrent rule: for \(\alpha>0\) and \(\Delta t_k>0\), the retention factor is contractive, while the second term injects the current stimulus. The cell emits an output \(y_k\) through a readout of the updated latent state and the current input,
\begin{equation}
y_k=c_o v_k+d_o u_k,
\end{equation}
which is passed to the next stage instead of exposing the latent state directly.

Cascading KNCs increases the effective order of the induced input--output dynamics. In the continuous-time reference form, eliminating the internal latent potential of the \(\ell\)-th stage yields
\begin{equation}
\left(C_\ell \frac{d}{dt}+a_\ell\right)y_\ell(t)
=
\left(d_{o,\ell}C_\ell\frac{d}{dt}+d_{o,\ell}a_\ell+c_{o,\ell}b_\ell\right)y_{\ell-1}(t),
\quad \ell=1,\dots,n,
\label{eq:results_stage_recursion}
\end{equation}
with \(y_0(t)\equiv u(t)\). Each stage therefore contributes one first-order differential factor between successive emitted outputs. Recursive composition forms an \(n\)-stage state evolution, or equivalently an \(n\)-th-order denominator in the generic non-degenerate input--output representation, up to possible pole--zero cancellations~\cite{kailath1980linear,ogata2010modern}. Thus, high-order evolution in CHONN is produced by internal cascade composition of Kirchhoff-governed cells rather than by manually appended positional or depth heuristics. The full end-to-end derivation is provided in Methods.

\begin{figure}[t]
\centering
\includegraphics[width=1.0\textwidth]{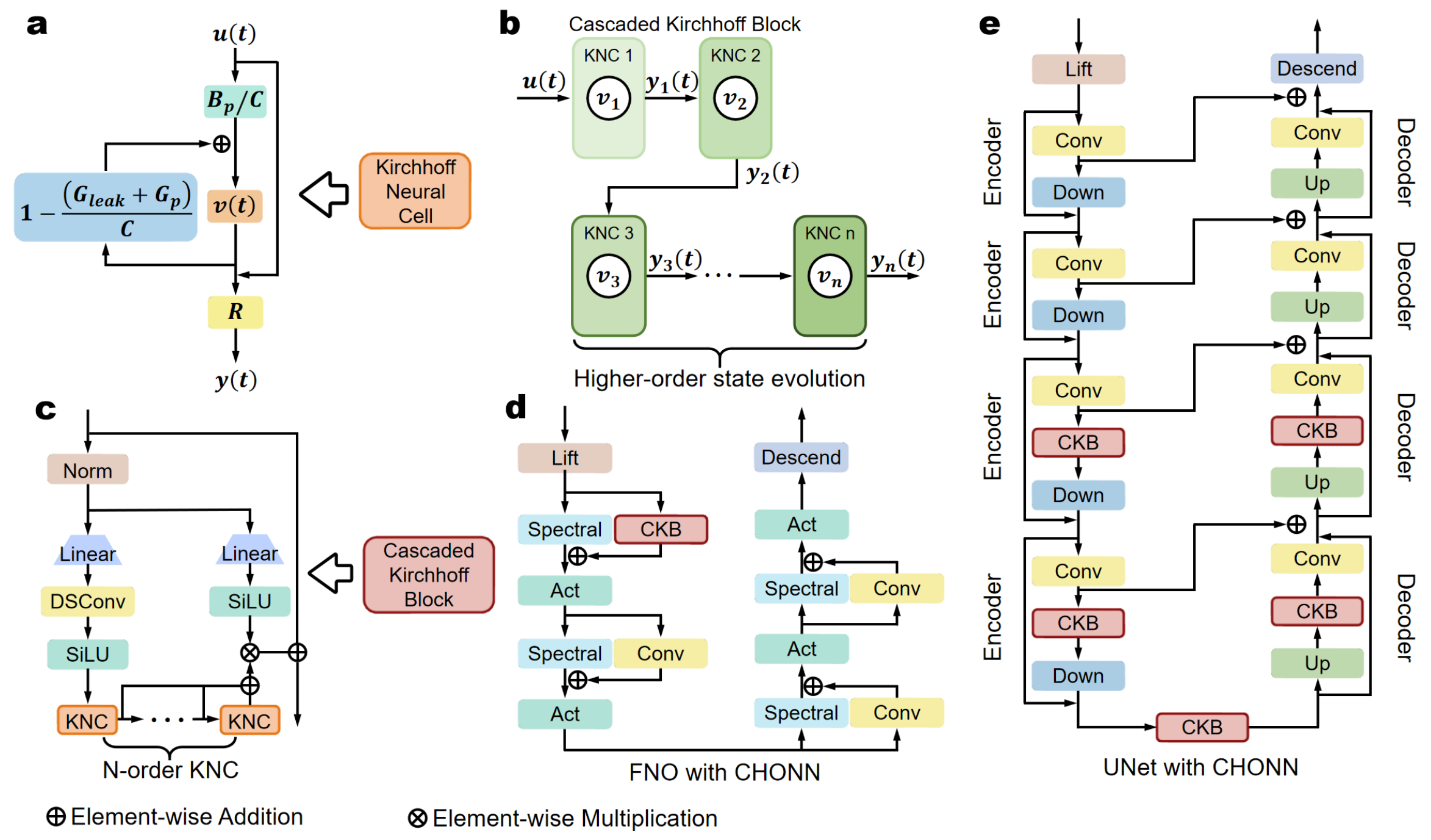}
\caption{
\textbf{Architectural instantiation of CHONN and its integration into neural-operator and encoder--decoder backbones.}
\textbf{a,} Structure of a single \emph{Kirchhoff Neural Cell} (KNC). The hidden representation is modeled as a latent potential \(v(t)\), where input injection, leakage dissipation and output readout jointly define a first-order latent evolution unit.
\textbf{b,} \emph{Cascaded Kirchhoff Block} (CKB). Multiple KNCs are connected in series, so that the output of each stage drives the next stage and progressively enriches the effective order of the latent dynamics.
\textbf{c,} Trainable realization of an \(N\)-stage CKB. The serial KNC cascade is embedded within a residual gated block together with normalization, projection, local mixing and nonlinear activation, making the high-order evolution module compatible with deep neural architectures.
\textbf{d,} \emph{FNO with CHONN}. CKB modules are inserted alongside spectral and convolutional transformations in the Fourier neural operator backbone, introducing explicit latent evolution into operator learning.
\textbf{e,} \emph{U-Net with CHONN}. CKB modules are placed within the encoder--decoder hierarchy, including deep downsampling, bottleneck and upsampling stages, to enrich multi-scale feature evolution.
Together, these instantiations show that CHONN functions as a modular plug-in mechanism for introducing intrinsic high-order representation evolution into different backbone architectures.
}
\label{fig2:model}
\end{figure}

\subsection*{CHONN realizes high-order representation evolution as a modular neural block}
\label{subsec:results_ckb_impl}

Fig.~\ref{fig2:model} shows how the Kirchhoff formulation is instantiated as a trainable neural block. At the unit level, a KNC implements the relaxation--injection--readout update derived above, using a latent potential as the internal evolution state and an emitted output for subsequent computation (Fig.~\ref{fig2:model}a). This preserves the interpretation of Kirchhoff-governed state evolution while allowing the effective coefficients to adapt to input features.

At the block level, CHONN constructs higher-order representation evolution by cascading multiple KNCs within a Cascaded Kirchhoff Block (CKB) (Fig.~\ref{fig2:model}b,c). Using \(r\) to index cascade stages, the serial composition is
\begin{equation}
y^{(1)}=\mathcal{K}_{\theta_1}(u),\qquad
y^{(r)}=\mathcal{K}_{\theta_r}\!\left(y^{(r-1)}\right),
\quad r=2,\dots,N .
\label{eq:results_knc_cascade}
\end{equation}
Each KNC contributes one circuit-inspired first-order evolution factor, and their serial composition forms the neural counterpart of the higher-order operator relation derived in Methods.

To retain responses from different evolutionary depths, CKB aggregates all stage outputs and fuses them with an input-conditioned gate:
\begin{equation}
\bar y=\sum_{r=1}^{N}y^{(r)},\qquad
Y=X+\bar y\odot g .
\label{eq:results_ckb_out_revised}
\end{equation}
Thus, the residual path preserves the input feature, while the cascaded branch contributes aggregated higher-order evolution responses. As illustrated in Fig.~\ref{fig2:model}d,e, the same CKB module can be inserted into both FNO-style neural operators and U-Net-style encoder--decoder backbones, showing that CHONN provides a modular mechanism for introducing intrinsic high-order representation evolution across architectures.

\begin{figure}[!ht]
\centering
\includegraphics[width=1.0\textwidth]{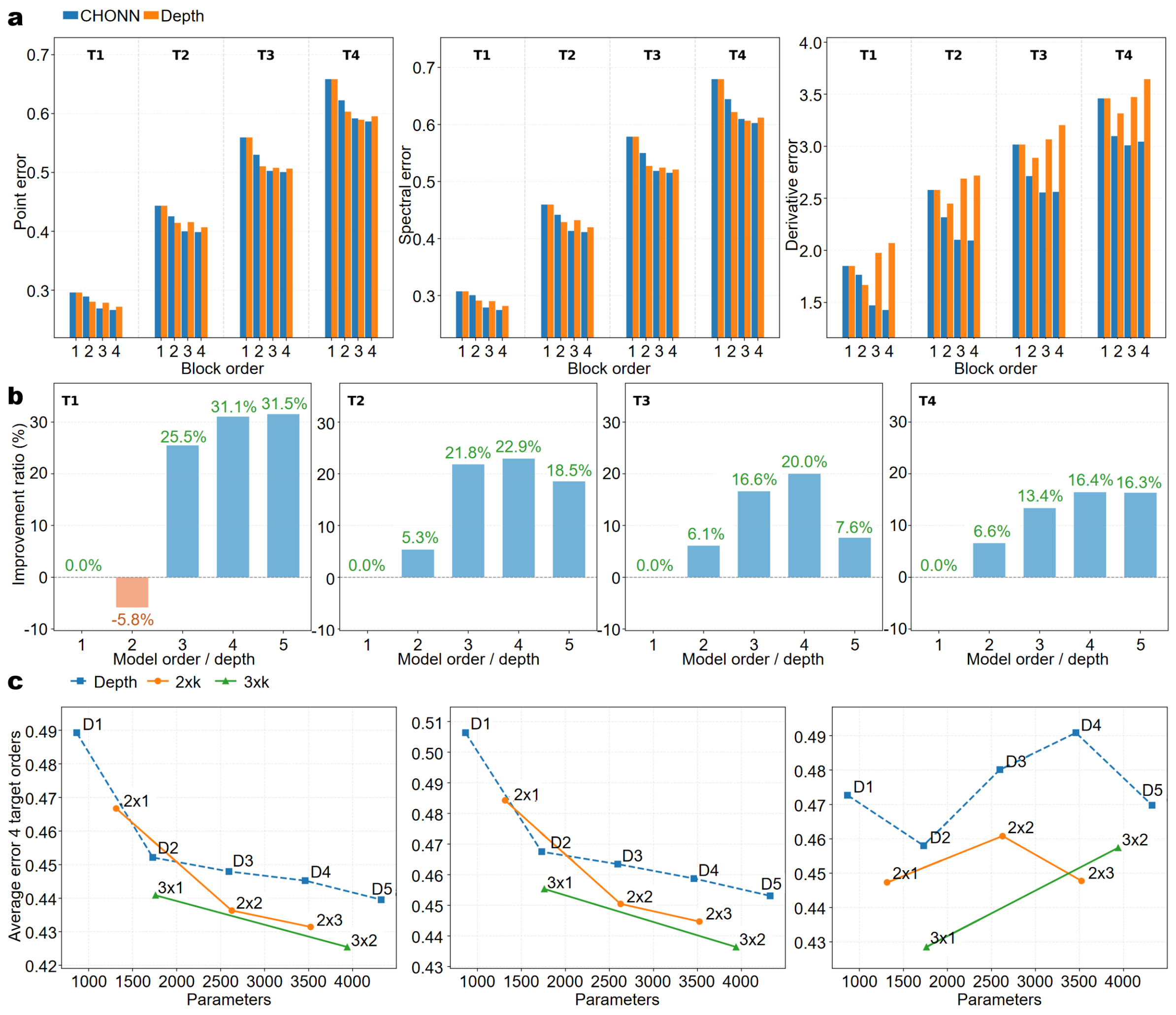}
\caption{
\textbf{Explicit high-order construction improves structural fidelity and changes the scaling path.}
\textbf{a,} Comparison between CHONN and a depth-stacked first-order baseline on the order-controlled benchmark. Target operators T1--T4 correspond to first- to fourth-order instances of the operator family. For each target order, physical-domain, spectral-domain and derivative relative \(L_2\) errors are reported across model orders or baseline depths 1--4. CHONN shows more consistent improvement with increasing order, with the clearest separation from the depth-stacked baseline appearing in the derivative-sensitive metric.
\textbf{b,} Relative reduction in derivative error achieved by CHONN over pure first-order depth stacking. Larger reductions are observed for higher target orders and stronger model configurations, indicating that explicit intra-block high-order construction is not equivalent to simply increasing the number of first-order blocks.
\textbf{c,} Parameter--performance scaling comparison between pure first-order depth scaling (D1--D5) and stacked high-order CHONN blocks (\(2\times k\) and \(3\times k\)). High-order block stacking follows a more favourable parameter--error trajectory, particularly for derivative-sensitive errors, suggesting that first increasing the order within each block and then stacking such blocks is more effective than continuing to deepen a purely first-order network.
}
\label{fig3:anal}
\end{figure}

\subsection*{Explicit intra-block high-order design beyond depth stacking}
\label{subsec:results_1d_highorder_vs_depth}

We validate whether CHONN's advantage arises from explicit intra-block high-order construction rather than from increasing network depth alone. We compare two designs under the same order-controlled benchmark: an \(n\)-order CKB, where \(n\) first-order KNCs are cascaded within a single block, and a depth-stacked baseline composed of repeated first-order CKBs~\cite{gu2023mamba}. The target tasks correspond to first- through fourth-order instances of
\begin{equation}
\left(I-\tau^2\partial_{ss}\right)^n y=x,
\end{equation}
denoted as T1--T4. Performance is evaluated using pointwise, spectral and derivative relative \(L_2\) errors, with the derivative metric serving as the most diagnostic measure of high-order structural fidelity.

As shown in Fig.~\ref{fig3:anal}a, increasing the CHONN block order produces a more stable improvement trend than depth stacking. The gains are visible in pointwise and spectral errors, but are most pronounced in derivative-sensitive error, indicating that intra-block high-order construction improves not only coarse reconstruction but also local structural preservation. By contrast, pure depth stacking improves coarse metrics but does not provide a stable improvement on the derivative metric; its derivative error often reaches the best value at a shallow depth and then degrades as depth continues to increase.

The matched-order comparison further separates the two mechanisms. For T1 the models are identical by construction, whereas for higher-order targets CHONN increasingly outperforms the depth-stacked baseline on derivative-sensitive errors (Fig.~\ref{fig3:anal}a,b). This trend shows that the advantage of CHONN becomes clearer when the target operator contains stronger high-order structure. Thus, explicit intra-block KNC cascading is not interchangeable with generic depth expansion; it implements a genuine high-order dynamical mechanism rather than a depth-induced approximation effect.

\subsection*{High-order block stacking provides a more favorable scaling path than pure first-order depth scaling}
\label{subsec:results_highorder_block_scaling}

We further compare two scaling strategies beyond a single block: continuing to deepen first-order blocks, or first constructing high-order dynamics within each block and then stacking such blocks. We analyze composite CHONN variants denoted by \(2\times2\), \(2\times3\), and \(3\times2\), and compare them with pure first-order depth-stacked baselines under similar parameter budgets (Fig.~\ref{fig3:anal}c).

Stacking high-order blocks yields a more favorable parameter--performance trajectory than pure first-order depth scaling. The advantage is consistent across pointwise, spectral and derivative-sensitive errors, and is strongest on the derivative metric, especially for higher-order targets. Under approximately matched budgets, \(2\times2\) improves over depth-3 and \(2\times3\) improves over depth-4, with the largest relative gains appearing in derivative-sensitive error. These results suggest a two-level compositional benefit: serial KNCs first construct higher-order recurrent operators within each block, and stacking these high-order blocks then further increases representational capacity. Therefore, making each block high-order before stacking provides a more parameter-efficient and structurally faithful scaling path than simply deepening a purely first-order network.

\begin{figure}[!ht]
\centering
\includegraphics[width=1.0\linewidth]{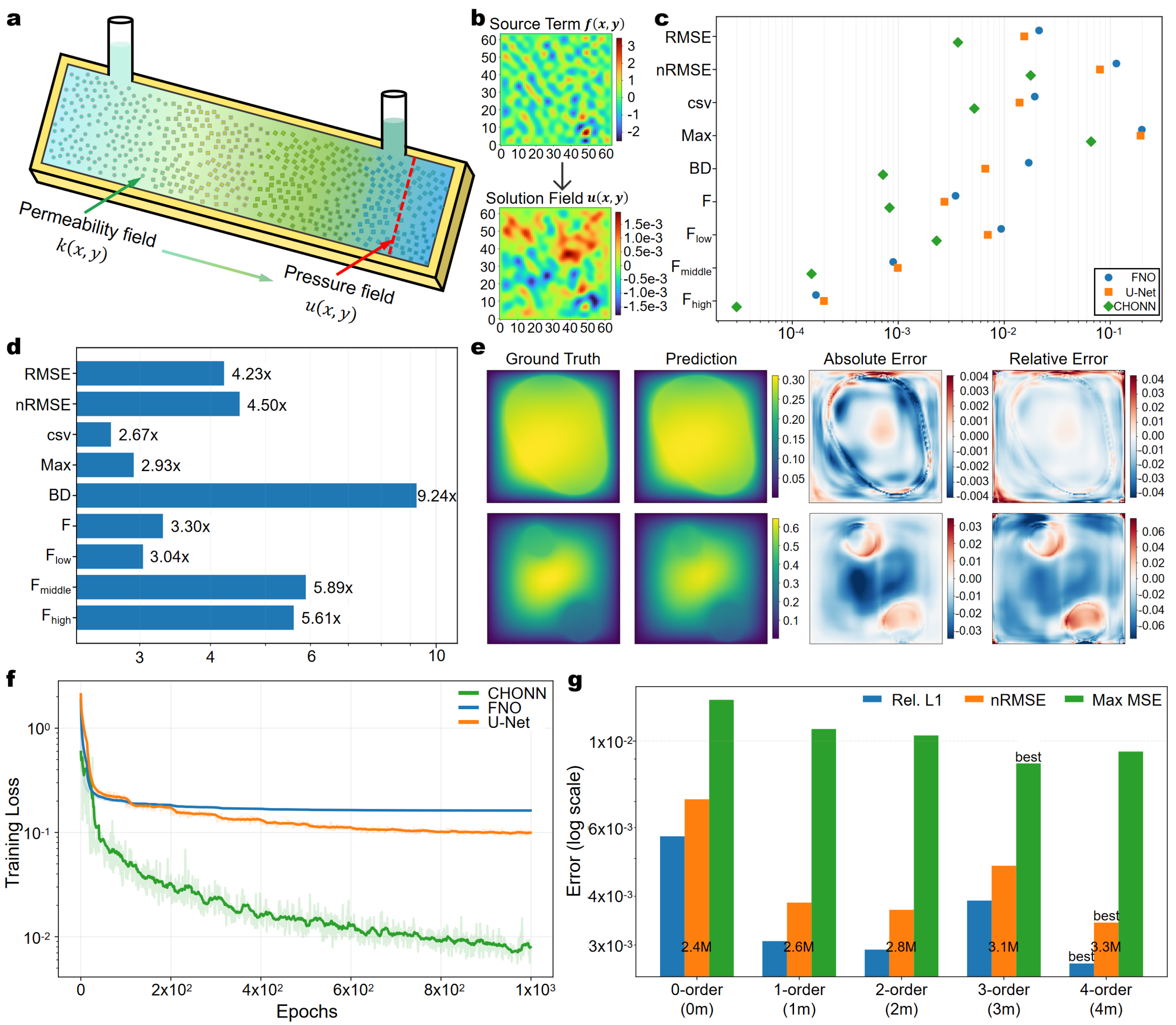}
\caption{
\textbf{CHONN improves steady-state neural operator learning on Darcy flow and Poisson problems.}
\textbf{a,} Physical illustration of the Darcy-flow problem, where a spatially heterogeneous permeability field \(k(x,y)\) determines the pressure field \(u(x,y)\) and the resulting flow through the porous medium.
\textbf{b,} Representative source term \(f(x,y)\) and solution field \(u(x,y)\) for the Poisson equation, illustrating the steady-state mapping from a prescribed spatial forcing field to the corresponding solution.
\textbf{c,} Multi-metric quantitative comparison on the Darcy benchmark. Across RMSE, nRMSE, CSV, maximum error, boundary discrepancy, and frequency-band errors, CHONN consistently achieves lower error than FNO and U-Net.
\textbf{d,} Relative gain of CHONN over the stronger Darcy baseline for each metric. The largest improvements appear in boundary-sensitive and frequency-related measures, indicating that the proposed model improves not only global accuracy but also structural fidelity.
\textbf{e,} Qualitative Darcy predictions and residual maps on representative samples. CHONN better preserves the global pressure structure and reduces localized artifacts, yielding cleaner absolute- and relative-error patterns than baseline models.
\textbf{f,} Training loss curves on the Darcy task. CHONN converges more rapidly and reaches a substantially lower loss floor, suggesting that the Kirchhoff-inspired representation leads to a more stable and efficient optimization process.
\textbf{g,} Cascade-stage ablation on the Poisson equation. Increasing the number of passes improves overall solution-field accuracy with only moderate growth in model size, while varying cascade depths yield distinct trade-offs among relative \(L_1\) error, nRMSE, and Max MSE.
Together, these results show that circuit-inspired high-order construction improves prediction accuracy, structural consistency, optimization stability, and scaling efficiency across representative steady-state PDE benchmarks.
}
\label{fig4:physics}
\end{figure}

\subsection*{Empirical validation across PDE benchmarks}\label{subsec:pde}

\subsubsection*{Darcy Flow Prediction with CHONN}
\label{subsubsec:darcy}

{Recent operator-learning studies have shown the promise of neural networks for learning PDE solution operators and nonlocal integral operators from data~\cite{goswami2022deep,zappala2024learning}.} {We first evaluate CHONN on Darcy flow, a classical steady-state elliptic PDE that maps a spatially heterogeneous permeability field \(a(x)\) to a pressure field \(u(x)\) (Fig.~\ref{fig4:physics}a)~\cite{evans2010partial,li2021fourier,takamoto2022pdebench}. 
On the domain \(D=[0,1]^2\), the solution is governed by
\begin{equation}
-\nabla \cdot \big(a(x)\nabla u(x)\big)=f(x), 
\quad x\in D,
\qquad 
u(x)=0,\quad x\in \partial D ,
\label{eq:darcy_main}
\end{equation}
where \(a(x)\) controls the spatially varying diffusion or permeability, \(f(x)\) is the forcing term, and \(u(x)\) is the corresponding steady-state response. 
Because the divergence-form elliptic operator couples local gradients with the global coefficient configuration and boundary constraints, Darcy flow provides a well-defined benchmark for evaluating whether CHONN can capture nonlocal spatial coupling and higher-order physical structure.}

We implement CHONN by inserting 2-order Cascaded Kirchhoff Blocks (CKBs) into a U-Net--based neural operator backbone, following the architectural design in Fig.~\ref{fig2:model}e. The CKB modules are placed in the deepest encoder--decoder stages, where the receptive fields are largest, and the latent features encode macro-scale spatial structures. This design allows the model to combine hierarchical multi-scale features with explicit high-order latent evolution.

CHONN consistently outperforms FNO and U-Net across the evaluated Darcy metrics (Fig.~\ref{fig4:physics}c). The advantage is visible not only in global errors such as RMSE and nRMSE, but also in local-fidelity metrics, including maximum error and frequency-resolved errors. The improvement ratios in Fig.~\ref{fig4:physics}d further show that CHONN provides the largest gains on metrics that are sensitive to boundary behavior and high-frequency spatial components, suggesting that the cascaded Kirchhoff evolution improves the recovery of fine-scale physical structures.

Visual comparisons support the same conclusion. As shown in Fig.~\ref{fig4:physics}e, CHONN produces pressure fields that closely match the ground truth while maintaining lower absolute and relative errors in high-gradient regions. In contrast, baseline predictions tend to exhibit blurred structures or localized artifacts near sharp transitions. The training curves in Fig.~\ref{fig4:physics}f further indicate that CHONN achieves a lower validation error than FNO and U-Net, suggesting that the proposed high-order evolution block improves both the optimization behavior and the final predictive accuracy.

Together, these results show that CHONN is effective for steady-state operator learning: it improves global prediction accuracy, reduces local errors, and better preserves sharp spatial structures in Darcy flow.
\subsubsection*{CKB order ablation on Poisson problems}
\label{subsubsec:poisson_ablation}

{We further evaluate whether increasing the cascade order of CKB improves the modeling of higher-order spatial operators using the Poisson equation~\cite{raonic2023convolutional,zheng2024alias}. 
Given a source field \(f(x,y)\) on the domain \(D=[0,1]^2\), the target solution \(u(x,y)\) is governed by
\begin{equation}
-\Delta u = f, \quad \text{in } D, 
\qquad 
u|_{\partial D}=0,
\label{eq:poisson_main}
\end{equation}
where \(\Delta=\partial_{xx}+\partial_{yy}\) is the second-order Laplace operator and homogeneous Dirichlet boundary conditions are imposed on \(\partial D\). 
The learning objective is therefore to approximate the solution operator as
\begin{equation}
\mathcal{G}^{\dagger}: f \mapsto u,
\label{eq:poisson_operator_main}
\end{equation}
which maps an input forcing field to its corresponding steady-state solution field, as illustrated in Fig.~\ref{fig4:physics}b. 
Because the solution is determined by the inverse action of a second-order elliptic operator, this benchmark provides a direct and controlled testbed for assessing whether cascaded KNCs can strengthen the approximation of globally coupled and higher-order spatial dependencies.}

For this ablation, we vary only the number of KNCs within the CKB module while keeping the backbone, training settings, and hyperparameters unchanged. We compare the original U-Net backbone without CKB modules as the 0-order baseline, and evaluate CKB variants from 1st-order to 4th-order.

The cascade-order ablation in Fig.~\ref{fig4:physics}g shows that increasing the number of orders leads to clear overall gains in Poisson solution prediction. Compared with the 0-order baseline, CKB variants reduce both relative \(L_1\) error and nRMSE, with the 4-order model achieving the best overall accuracy. These results indicate that higher-order cascaded evolution improves global solution-field modeling.

The error trends also suggest that different cascade orders emphasize complementary aspects of the solution. Moderate orders already improve average prediction accuracy, while higher orders further strengthen the recovery of localized high-gradient structures and reduce global errors.

Together, these results support the theoretical formulation that cascading KNCs increases the effective order of the operator realized by the network. In the Poisson setting, higher-order CKB variants provide stronger solution-field modeling capacity, with the best trade-off achieved by the 4-order configuration.

\begin{figure}[!ht]
\centering
\includegraphics[width=1.0\linewidth]{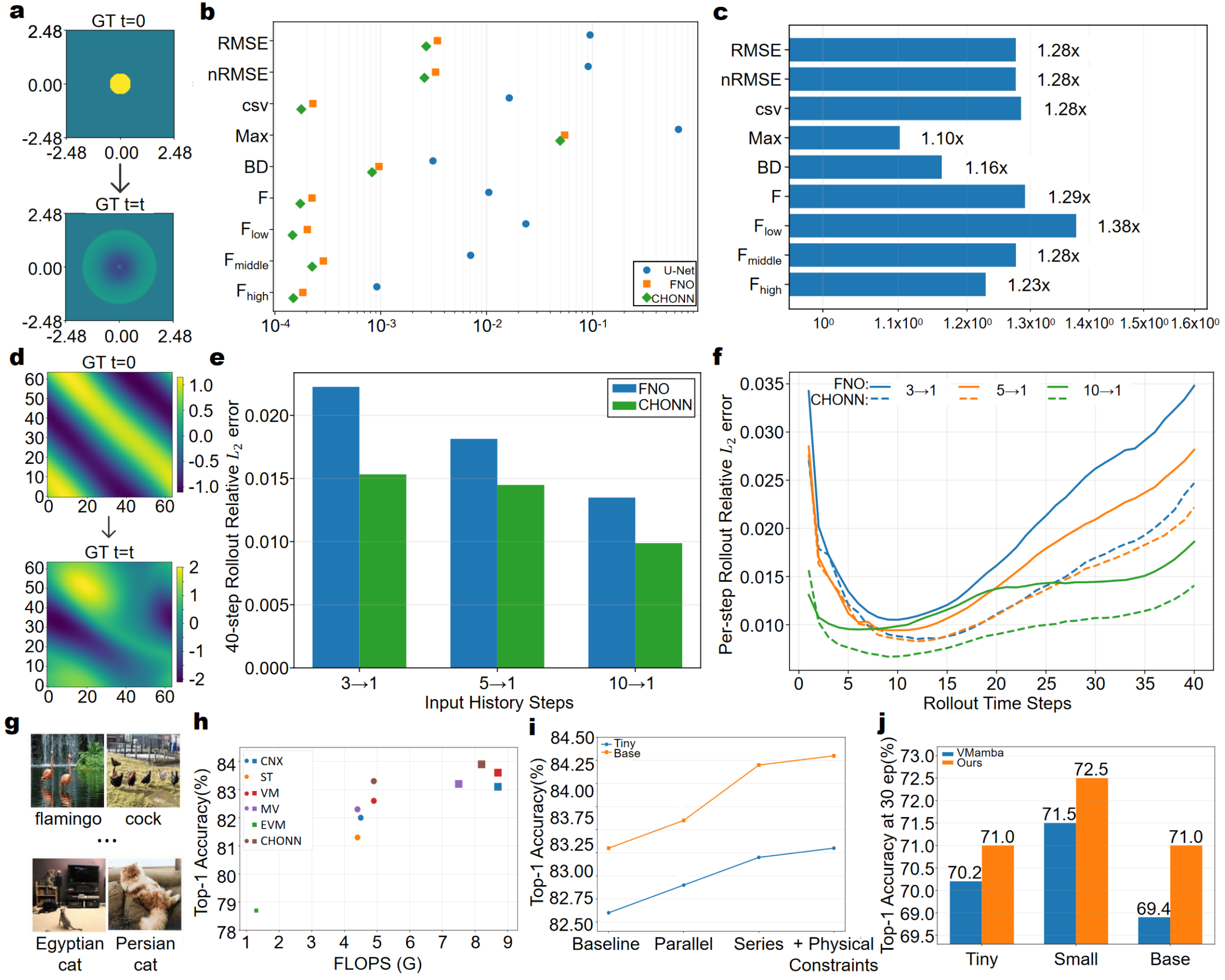}
\caption{
\textbf{Circuit-inspired high-order construction improves spatiotemporal forecasting, long-horizon rollout stability and visual recognition.}
\textbf{a,} Representative shallow-water evolution from an initial localized height disturbance at \(t=0\) to the propagated wave field at a later time \(t\), illustrating the spatiotemporal forecasting task.
\textbf{b,} Multi-metric comparison on the shallow-water benchmark. CHONN is compared with U-Net and FNO across RMSE, nRMSE, CSV, maximum error, boundary discrepancy and frequency-related errors, including low-, middle- and high-frequency components.
\textbf{c,} Relative gain of CHONN over the stronger shallow-water baseline for each metric, showing consistent improvements in global reconstruction, boundary consistency and frequency-sensitive structural fidelity.
\textbf{d,} Representative Navier--Stokes vorticity fields at the beginning and end of a rollout sequence, showing the transition from an initial field to a more complex flow structure.
\textbf{e,} Average 40-step rollout relative \(L_2\) error on Navier--Stokes dynamics under different input-history settings (\(3\!\to\!1\), \(5\!\to\!1\) and \(10\!\to\!1\)). CHONN consistently yields lower rollout error than FNO.
\textbf{f,} Per-step rollout error curves for FNO and CHONN under different input-history settings. CHONN shows lower error accumulation over the rollout horizon, with a clearer advantage in later prediction steps.
\textbf{g,} Representative natural-image samples from ImageNet-1K.
\textbf{h,} Accuracy--complexity comparison on ImageNet-1K, with Top-1 accuracy plotted against FLOPs for representative vision backbones. CHONN reaches a favourable accuracy--complexity trade-off compared with ConvNeXt (CNX), Swin Transformer (ST), VMamba (VM), MambaVision (MV) and EfficientVMamba (EVM).
\textbf{i,} Ablation of routing topology and circuit constraints. Series cascade yields larger gains than parallel expansion, and adding Kirchhoff-inspired constraints further improves accuracy.
\textbf{j,} Training efficiency at 30 epochs. CHONN achieves higher early-stage Top-1 accuracy than VMamba across Tiny, Small and Base settings.
Together, these results show that Kirchhoff-inspired high-order dynamics improve spatiotemporal field prediction, long-horizon rollout stability and visual representation learning across different task families.
}
\label{fig5:ns}
\end{figure}
\subsubsection*{Shallow-Water Forecasting with CHONN}
\label{subsubsec:sw}

{To evaluate CHONN's ability to model non-stationary spatiotemporal evolution, we apply our method to the two-dimensional Shallow Water equations (SWE) dataset~\cite{takamoto2022pdebench}. 
The dynamics are governed by the conservation of water mass and momentum:
\begin{equation}
\begin{aligned}
\partial_t h + \partial_x(hu) + \partial_y(hv) &= 0, \\
\partial_t(hu) + \partial_x\left(u^2h + \frac{1}{2}g_r h^2\right) + \partial_y(uvh) &= -g_r h\,\partial_x b, \\
\partial_t(hv) + \partial_y\left(v^2h + \frac{1}{2}g_r h^2\right) + \partial_x(uvh) &= -g_r h\,\partial_y b .
\end{aligned}
\label{eq:shallow_water}
\end{equation}
where \(h\) denotes the water depth, \(u\) and \(v\) are the horizontal velocity components, \(b\) is the bathymetry, and \(g_r\) is the gravitational acceleration.}

We implement CHONN using a modified Fourier neural operator backbone~\cite{li2021fourier} with 2-order CKBs, as described in Fig.~\ref{fig2:model}d. The model takes previous temporal snapshots and spatial coordinates as input, lifts them into a latent representation, and uses the proposed CKB module to introduce Kirchhoff-inspired high-order latent evolution.

Fig.~\ref{fig5:ns}b reports a multi-metric comparison among U-Net~\cite{ronneberger2015unet}, FNO~\cite{li2021fourier}, and CHONN on the SWE benchmark. CHONN consistently achieves lower errors across global reconstruction, boundary discrepancy, and frequency-band measures, indicating improved field prediction and structure-sensitive fidelity.

The relative-gain analysis in Fig.~\ref{fig5:ns}c further shows that CHONN improves over the stronger baseline across all evaluated metrics. The gains are observed not only in RMSE-based measures, but also in boundary-sensitive and frequency-dependent errors, suggesting that the proposed high-order construction better preserves multi-scale spatial structure during temporal evolution.

Together, these results show that the Kirchhoff-inspired high-order construction improves SWE forecasting in both global accuracy and structure-sensitive measures.

\subsubsection*{Rollout Horizon Evaluation on Navier--Stokes Dynamics}
\label{subsubsec:ns}

{To further evaluate temporal stability and rollout robustness, we conduct long-horizon forecasting experiments on the two-dimensional incompressible Navier--Stokes equations in the vorticity formulation with viscosity \(\nu=10^{-3}\), following the standard neural-operator benchmark setting~\cite{li2021fourier}. 
The dynamics are governed by
\begin{equation}
\partial_t \omega(x,t) + u(x,t)\cdot\nabla \omega(x,t)
=
\nu \Delta \omega(x,t) + f(x),
\qquad x\in (0,1)^2,\ t\in(0,T],
\label{eq:ns_vorticity_main}
\end{equation}
with the incompressibility constraint as 
\begin{equation}
\nabla\cdot u(x,t)=0,
\label{eq:ns_incompressible_main}
\end{equation}
where \(\omega=\nabla\times u\) denotes the vorticity field, \(u\) is the velocity field, and \(f(x)\) is a fixed forcing term. 
This benchmark couples nonlinear advection, viscous diffusion and multi-scale vortex interactions, making recursive prediction sensitive to small local errors that can accumulate over time~\cite{temam2001navier}. 
As illustrated in Fig.~\ref{fig5:ns}d, the vorticity field evolves from an initial spatial pattern into increasingly complex flow structures, requiring the model to preserve coherent dynamics throughout long rollouts.}

We compare CHONN with FNO under different input-history settings, denoted as \(3\!\to\!1\), \(5\!\to\!1\), and \(10\!\to\!1\). Here, \(m\!\to\!1\) means that the model takes the previous \(m\) vorticity frames as input and predicts the next one-step field. During evaluation, this one-step predictor is applied recursively to generate a 40-step rollout. We report the average rollout relative \(L_2\) error and the corresponding per-step error trajectories in Fig.~\ref{fig5:ns}e,f.

As shown in Fig.~\ref{fig5:ns}e, CHONN consistently achieves lower 40-step rollout error than FNO across all evaluated input-history settings. The advantage holds for both short and relatively longer temporal contexts, indicating that the proposed high-order circuit construction improves rollout accuracy beyond one-step prediction.

The comparison also shows that CHONN is less sensitive to shortened temporal context. When fewer historical frames are available, the performance degradation of CHONN is milder than that of FNO, suggesting that the cascaded circuit construction provides a stronger internal evolution prior and helps infer future dynamics from limited observations.

The per-step curves in Fig.~\ref{fig5:ns}f provide a more detailed view of the rollout process. Across all three input-history settings, CHONN remains consistently below FNO through most of the 40-step rollout. The gap becomes especially visible in later prediction steps, where FNO accumulates error more rapidly, whereas CHONN maintains a flatter trajectory.

Together, Fig.~\ref{fig5:ns}d--f show that the proposed high-order circuit construction improves long-horizon Navier--Stokes forecasting. It reduces average rollout error across different temporal contexts, degrades more mildly when the input history is shortened, and slows the accumulation of prediction error during recursive temporal evolution.

\subsection*{Generalization to large-scale natural image recognition}
\label{subsec:vision}

To examine whether the proposed circuit-inspired high-order construction also benefits general visual representation learning, we evaluate CHONN on the ImageNet-1K classification benchmark~\cite{deng2009imagenet} (Fig.~\ref{fig5:ns}g). Although this setting differs from PDE forecasting, it still requires hierarchical transformation of complex two-dimensional spatial representations.

As summarized in Fig.~\ref{fig5:ns}h, CHONN achieves a favorable accuracy--complexity trade-off compared with representative convolutional, transformer-based, and visual state-space baselines, including ConvNeXt~\cite{liu2022convnet}, Swin Transformer~\cite{liu2021swin}, VMamba~\cite{liu2024vmamba}, MambaVision~\cite{hatamizadeh2025mambavision}, and EfficientVMamba~\cite{pei2025efficientvmamba}. Across model scales, CHONN consistently improves recognition accuracy while maintaining competitive parameter and computational costs, indicating that the proposed module is not limited to explicit PDE settings.

We further analyze the source of this improvement through the routing and constraint ablation in Fig.~\ref{fig5:ns}i. Parallel expansion provides moderate gains, whereas series cascade brings a larger improvement, and adding explicit Kirchhoff constraints further improves accuracy. This comparison supports the design choice that sequential high-order composition is more effective than parallel branch expansion for modeling representation dynamics.

Finally, Fig.~\ref{fig5:ns}j compares training efficiency at 30 epochs. Across Tiny, Small, and Base settings, CHONN achieves higher early-stage accuracy than VMamba, suggesting that the circuit-inspired high-order construction not only improves final recognition performance but also provides a more effective optimization trajectory.

Together, Fig.~\ref{fig5:ns}g--j show that CHONN generalizes beyond scientific operator learning: the same high-order circuit-inspired mechanism improves large-scale visual recognition, benefits from cascaded routing, and accelerates early-stage training. 
\section*{Discussion}

This work presents CHONN as a circuit-inspired framework for modeling representation evolution through intrinsic latent dynamics. Rather than treating evolution as an external positional cue or as an implicit consequence of depth, CHONN formulates each representation update as a Kirchhoff-governed latent-potential process and then raises its effective order through cascade composition. In this formulation, a single Kirchhoff Neural Cell provides a stable first-order evolution primitive, while the Cascaded Kirchhoff Block composes multiple such primitives to construct higher-order dynamics within a single representational layer. This perspective provides a mechanistic link between circuit-inspired state evolution, neural architecture design, and task-level behavior.

The empirical results support this interpretation at two complementary levels. On the order-controlled benchmark, explicit intra-block high-order construction behaves differently from generic stacking of first-order blocks. The advantage is most visible on derivative-sensitive errors, where high-order structural mismatch is more directly exposed than in coarse pointwise reconstruction. This observation is important because it indicates that the benefit of CHONN does not simply arise from increased parameter count or greater network depth. Instead, making the block itself high-order changes the form of the realized dynamics and leads to a more favorable scaling path when such blocks are further stacked.

Across PDE benchmarks, the same mechanism translates into improved structural fidelity. In steady-state Darcy and Poisson problems, CHONN better preserves globally coupled solution fields, boundary-sensitive behavior and localized high-gradient structures. In time-dependent shallow-water and Navier--Stokes settings, the model reduces rollout error and slows error accumulation over long recursive predictions. These results suggest that circuit-inspired high-order composition is particularly useful when the target mapping depends on spatial derivatives, multi-scale interactions or temporally accumulated dynamics. The improvements are therefore consistent with the central claim that representation evolution should be modeled as an internal dynamical process, rather than appended as an external architectural device.

The ImageNet-1K results further indicate that the proposed construction is not restricted to explicit physical simulation. Although natural-image recognition does not provide a known governing equation in the same way as PDE benchmarks, it still requires hierarchical transformation of spatial representations. The accuracy--complexity comparison and routing ablations show that serial high-order composition can also improve visual representation learning, especially when combined with explicit circuit constraints. This does not imply that visual recognition is governed by the same physical dynamics as the PDE systems studied here. Rather, it suggests that high-order latent evolution can serve as a useful architectural prior beyond scientific machine learning.

Several limitations remain. First, the appropriate cascade order is currently selected empirically. Although the controlled benchmark and Poisson ablation show clear benefits from increasing order, the optimal order may depend on the target operator, data resolution, backbone architecture and computational budget. Developing adaptive mechanisms that select or regularize the effective order during training would make the framework more flexible. Second, CHONN introduces an additional structural prior, but it does not replace the need for strong backbone design, careful training protocols or sufficient data coverage. Third, the current analysis focuses on deterministic supervised benchmarks; extending the framework to noisy observations, irregular domains, uncertainty-aware forecasting and real experimental measurements would further test its robustness.

Overall, these findings establish high-order circuit composition as a principled and practical route for neural dynamics modeling. By grounding representation updates in Kirchhoff-inspired latent evolution and lifting first-order cells through cascade composition, CHONN provides an interpretable mechanism for enriching the dynamical order of neural representations. The resulting gains across controlled operators, physical forecasting and large-scale visual recognition suggest that explicit high-order evolution is a useful design principle for models that must preserve structure, propagate information stably and scale beyond shallow first-order transformations.
\section*{Methods}

\paragraph{Mathematical formulation of the Kirchhoff Neural Cell.}

We describe representation learning through a Kirchhoff-governed latent potential \(v(t)\), which serves as the cell-internal evolution variable. For a single Kirchhoff cell driven by an external input \(u(t)\), Kirchhoff's current law gives
\begin{equation}
i_C(t) + i_{\text{leak}}(t) + i_{\text{couple}}(t) = i_{\text{in}}(t),
\label{eq:methods_kcl_sum}
\end{equation}
where the branch currents are defined as
\begin{equation}
i_C(t)=C\frac{dv(t)}{dt},\qquad
i_{\text{leak}}(t)=G_{\text{leak}}\,v(t),\qquad
i_{\text{couple}}(t)=G_p\,v(t),\qquad
i_{\text{in}}(t)=B_p\,u(t).
\label{eq:methods_currents_def}
\end{equation}
Substituting Eq.~\eqref{eq:methods_currents_def} into Eq.~\eqref{eq:methods_kcl_sum} yields
\begin{equation}
C\frac{dv(t)}{dt}
=
-(G_{\text{leak}}+G_p)\,v(t)+B_p\,u(t).
\label{eq:methods_rc_ode}
\end{equation}
We refer to Eq.~\eqref{eq:methods_rc_ode} as the \emph{Kirchhoff representation} in this work: it specifies how a soma-level latent potential evolves under Kirchhoff-consistent relaxation and input coupling, and it serves as the primitive that is recursively lifted to higher-order circuit operators.
Defining
\[
\alpha \triangleq \frac{G_{\text{leak}}+G_p}{C},
\qquad
\beta \triangleq \frac{B_p}{C},
\]
the dynamics can be written as
\begin{equation}
\dot v(t)= -\alpha\,v(t) + \beta\,u(t),
\label{eq:methods_state_ct}
\end{equation}
which makes explicit that the latent evolution is jointly governed by relaxation and input injection.

To deploy this continuous-time mechanism in neural computation, we discretize Eq.~\eqref{eq:methods_state_ct} under the zero-order hold (ZOH) assumption~\cite{hasani2022closed,higham2008functions}. Let \(t_k\) denote the \(k\)-th discretization point, \(\Delta t_k=t_k-t_{k-1}\), \(v_k=v(t_k)\), and \(u_k=u(t_k)\). Specifically, the input is assumed to be piecewise constant within each interval, i.e.,
\begin{equation}
u(t_{k-1}+\xi)=u_k,\qquad \xi\in[0,\Delta t_k].
\end{equation}
Under this assumption and fixed coefficients within the interval, the continuous-time dynamics admit the following closed-form update:
\begin{align}
v_k
&= e^{-\alpha\Delta t_k}v_{k-1}
+ \int_{0}^{\Delta t_k} e^{-\alpha(\Delta t_k-\xi)} \beta u_k\,d\xi \nonumber\\
&= e^{-\alpha\Delta t_k}v_{k-1}
+ \frac{\beta}{\alpha}\left(1-e^{-\alpha\Delta t_k}\right)u_k.
\label{eq:methods_zoh_exact}
\end{align}
Equivalently, by defining
\[
\bar{\alpha}_k \triangleq e^{-\alpha\Delta t_k},
\qquad
\bar{\beta}_k \triangleq
\frac{\beta}{\alpha}\left(1-e^{-\alpha\Delta t_k}\right),
\]
the discrete update can be written in the standard discrete state-space form
\begin{equation}
v_k=\bar{\alpha}_k v_{k-1}+\bar{\beta}_k u_k.
\label{eq:methods_state_discrete}
\end{equation}
When $\alpha>0$ and $\Delta t_k>0$, the retention factor satisfies $0<e^{-\alpha\Delta t_k}<1$. Thus, the discrete update gives a contractive latent retention term together with an input-injection term.

A KNC does not expose its internal state directly. Instead, its output is read out from both the updated state and the current input:
\begin{equation}
y_k = c_o\,v_k + d_o\,u_k,
\label{eq:methods_knc_output_discrete}
\end{equation}
where $y_k$ is the output voltage of the cell at the $k$-th discrete step, and $c_o$ and $d_o$ are fixed readout coefficients. This separates \emph{internal evolution} from \emph{external emission}: the internal state stores and transforms latent dynamics, whereas the output is the quantity propagated to the next cell.

For continuous-time analysis, we use the corresponding readout form as
\begin{equation}
y(t)=c_o\,v(t)+d_o\,u(t).
\label{eq:methods_knc_output_ct}
\end{equation}

\paragraph{High-order evolution via the Cascaded Kirchhoff Block.}
Let \(y_0(t) \equiv u(t)\) denote the external input in the continuous-time reference form. For the \(\ell\)-th cell in a cascade, we define the state-update and output-readout equations as
\begin{equation}
C_\ell \frac{d v_\ell(t)}{dt}
=
-a_\ell v_\ell(t) + b_\ell y_{\ell-1}(t),
\label{eq:methods_cascade_state}
\end{equation}
\begin{equation}
y_\ell(t)
=
c_{o,\ell} v_\ell(t) + d_{o,\ell} y_{\ell-1}(t),
\qquad \ell=1,\dots,n.
\label{eq:methods_cascade_output}
\end{equation}
Here, \(v_\ell(t)\) is the internal state of the \(\ell\)-th KNC, \(y_\ell(t)\) is its emitted output, and the next stage receives \(y_\ell(t)\) rather than \(v_\ell(t)\). This formalizes the intended KNC--CKB hierarchy: each cell first evolves its latent voltage state and then emits an output voltage that drives the next cell.

Assuming \(c_{o,\ell}\neq 0\), the internal state can be eliminated from Eq.~\eqref{eq:methods_cascade_output} as
\[
v_\ell(t)=\frac{y_\ell(t)-d_{o,\ell}y_{\ell-1}(t)}{c_{o,\ell}}.
\]
Substituting this into Eq.~\eqref{eq:methods_cascade_state} yields the output-level operator recursion
\begin{equation}
\left(C_\ell \frac{d}{dt}+a_\ell\right)y_\ell(t)
=
\left(
d_{o,\ell} C_\ell \frac{d}{dt}
+ d_{o,\ell} a_\ell
+ c_{o,\ell} b_\ell
\right)
y_{\ell-1}(t).
\label{eq:methods_single_stage_operator}
\end{equation}
Eq.~\eqref{eq:methods_single_stage_operator} shows that each stage induces a first-order operator relation between successive emitted outputs.

For clarity, applying this relation stage by stage gives the explicit recursive expansion
\begin{equation}
\begin{aligned}
\left(C_n \frac{d}{dt}+a_n\right)y_n(t)
&=
\left(
d_{o,n}C_n\frac{d}{dt}
+d_{o,n}a_n
+c_{o,n}b_n
\right)y_{n-1}(t),\\
\left(C_{n-1} \frac{d}{dt}+a_{n-1}\right)y_{n-1}(t)
&=
\left(
d_{o,n-1}C_{n-1}\frac{d}{dt}
+d_{o,n-1}a_{n-1}
+c_{o,n-1}b_{n-1}
\right)y_{n-2}(t),\\
&\;\;\vdots\\
\left(C_1 \frac{d}{dt}+a_1\right)y_1(t)
&=
\left(
d_{o,1}C_1\frac{d}{dt}
+d_{o,1}a_1
+c_{o,1}b_1
\right)y_0(t),
\end{aligned}
\label{eq:methods_recursive_operator_expanded}
\end{equation}
where \(y_0(t)\equiv u(t)\).

Under the fixed-coefficient linear reference setting, recursive application of Eq.~\eqref{eq:methods_single_stage_operator}, equivalently the expanded system in Eq.~\eqref{eq:methods_recursive_operator_expanded}, yields the end-to-end operator form
\begin{equation}
\prod_{\ell=1}^{n}
\left(C_\ell \frac{d}{dt}+a_\ell\right)
y_n(t)
=
\prod_{\ell=1}^{n}
\left(
d_{o,\ell} C_\ell \frac{d}{dt}
+ d_{o,\ell} a_\ell
+ c_{o,\ell} b_\ell
\right)
u(t).
\label{eq:methods_end_to_end_operator}
\end{equation}
Eq.~\eqref{eq:methods_end_to_end_operator} shows that, in the continuous-time linear reference form, the cascade produces a product of \(n\) first-order differential factors acting on the final-stage output \(y_n(t)\). In the generic non-degenerate case and up to possible pole--zero cancellations, this corresponds to an \(n\)-stage state evolution, or equivalently an \(n\)-th-order denominator in the associated input--output representation~\cite{kailath1980linear,ogata2010modern}. This formulation provides a reference model for interpreting how serially composed KNC stages increase the effective order of the induced input--output evolution.

The formulation above clarifies three levels of structure in CHONN. First, a single KNC implements a first-order state-evolution unit grounded in RC dynamics. Second, ZOH discretization turns these continuous dynamics into a closed-form recurrent update suitable for neural computation. Third, the CKB composes state-update/readout pairs across stages, providing an explicit intra-block mechanism for constructing higher-order latent evolution.
\paragraph{Neural architecture of Kirchhoff Neural Cell and Cascaded Kirchhoff Block.}
The theoretical formulation above defines a Kirchhoff Neural Cell (KNC) as the neural realization of the Kirchhoff representation, with three essential components: relaxation, input injection, and readout. In the neural implementation, we preserve the same evolution/readout structure but instantiate some of these coefficients in a selective, discrete form so that the effective dynamics can adapt to the current feature. This construction should be interpreted as a theory-to-realization discretization of the Kirchhoff representation, rather than as an ad hoc modification of a pre-existing engineering scan kernel. Let the input feature be
\begin{equation}
X \in \mathbb{R}^{B\times L\times d},
\label{eq:methods_ckb_input}
\end{equation}
where \(B\) is the batch size, \(L\) is the sequence length (or the flattened spatial length for 2D features), and \(d\) is the channel dimension. The block output is
\begin{equation}
Y \in \mathbb{R}^{B\times L\times d},
\label{eq:methods_ckb_output}
\end{equation}
with the same shape as the input, so that the block can be seamlessly inserted into deep hierarchical backbones.

Given the normalized input \(\widetilde X=\mathrm{Norm}(X)\), the block first splits into an evolution branch and a gate branch:
\begin{equation}
u = \phi\!\big(\mathrm{DSConv}(W_u \widetilde X)\big),
\qquad
g = \phi(W_g \widetilde X),
\label{eq:methods_dual_branch}
\end{equation}
where \(W_u\) and \(W_g\) are learnable linear projections, \(\mathrm{DSConv}(\cdot)\) denotes depthwise separable convolution, and \(\phi(\cdot)\) denotes the SiLU activation. The evolution branch provides the driving signal for Kirchhoff-governed latent evolution, while the gate branch provides an input-dependent modulation path. The DSConv layer introduces lightweight local interaction before the dynamical update.

For the discrete neural realization of a single KNC, we use \(v_k\) to denote the latent potential at the \(k\)-th discrete step, in direct correspondence with the continuous-time variable \(v(t)\) sampled at \(t_k\). The implemented evolution/readout rule is written as
\begin{equation}
v_k = \bar\alpha_k \odot v_{k-1} + \bar\beta_k \odot u_k,
\qquad
y_k = c_k \odot v_k + d\,u_k,
\label{eq:methods_knc_impl}
\end{equation}
where \(y_k\) is the emitted output of the cell at the \(k\)-th discrete step, \(\bar\alpha_k\) is the effective retention coefficient, \(\bar\beta_k\) is the effective input-injection coefficient, \(c_k\) is the readout coefficient, and \(d\) is a learned skip/readout parameter. This is the discrete implementation counterpart of Eqs.~\eqref{eq:methods_state_discrete} and \eqref{eq:methods_knc_output_discrete}: \(\bar\alpha_k\) corresponds to relaxation/retention, \(\bar\beta_k\) corresponds to input injection, and \((c_k,d)\) correspond to latent-potential/input readout.

To enhance expressive capacity, part of these effective coefficients are made input-dependent through learnable projections:
\begin{equation}
\Delta_k = \mathrm{softplus}(W_\Delta u_k + b_\Delta),\qquad
b_k = W_b u_k,\qquad
c_k = W_c u_k,
\label{eq:methods_selective_proj}
\end{equation}
and the discretized coefficients are formed as
\begin{equation}
\bar\alpha_k = \exp(-\Delta_k \odot \lambda),
\qquad
\bar\beta_k =
\frac{1-\exp(-\Delta_k \odot \lambda)}{\lambda}
\odot b_k,
\label{eq:methods_selective_disc}
\end{equation}
where \(\lambda\) is a learned decay parameter. In this parameterization, \(\lambda\) plays the role of the continuous-time decay rate, \(\Delta_k\) acts as an input-dependent effective discretization timescale, and the combination \(\exp(-\Delta_k\odot\lambda)\) is the selective implementation counterpart of the theoretical retention factor \(e^{-\alpha \Delta t_k}\) in Eq.~\eqref{eq:methods_zoh_exact}. Thus, the network implementation preserves the Kirchhoff-inspired relaxation--injection--readout interpretation while allowing the dynamics to adapt to the current feature.

The evolution branch is then processed by a cascaded Kirchhoff operator composed of serial KNCs. Denoting the \(r\)-th KNC by \(\mathcal{K}_{\theta_r}(\cdot)\), the cascade is written as
\begin{equation}
y^{(1)} = \mathcal{K}_{\theta_1}(u), \qquad
y^{(r)} = \mathcal{K}_{\theta_r}\!\big(y^{(r-1)}\big), \quad r=2,\dots,N,
\label{eq:methods_knc_cascade}
\end{equation}
where \(N\) is the cascade depth. This realizes the discrete counterpart of the higher-order cascade derived in Eqs.~\eqref{eq:methods_single_stage_operator}--\eqref{eq:methods_end_to_end_operator}: each KNC contributes a first-order Kirchhoff-governed evolution factor, and serial composition progressively enriches the effective order of the end-to-end operator.

To preserve shallow, intermediate, and deep evolutionary responses, we aggregate the outputs of all KNC stages into the final cascaded representation:
\begin{equation}
\bar y = \sum_{r=1}^{N} y^{(r)},
\label{eq:methods_cko_fuse}
\end{equation}
where \(y^{(r)}\) denotes the output of the \(r\)-th KNC in the cascade. Equivalently, this can be viewed as progressively adding all preceding-stage responses to the final-stage pathway. Such dense aggregation preserves low- and intermediate-order evolutionary cues while integrating the deepest response from the cascaded operator.

The gate branch then modulates the fused feature through element-wise multiplication:
\begin{equation}
\widehat y = \bar y \odot g,
\label{eq:methods_gate_fuse}
\end{equation}
and the final block output is obtained by residual addition:
\begin{equation}
Y = X + \widehat y.
\label{eq:methods_final_out}
\end{equation}

Therefore, the proposed CKB implements the theoretical KNC--CKB hierarchy in a practical neural form: a single KNC realizes selective Kirchhoff-governed latent evolution, while the full block combines local preprocessing, cascaded multi-order evolution, dense cross-stage aggregation, input-aware gating, and residual fusion within a unified architecture. Motivated by recent concerns that weak baselines and reporting biases can lead to overoptimistic conclusions in machine-learning-based PDE solving~\cite{mcgreivy2024weak}, we compare CHONN against established neural-operator and encoder--decoder baselines under matched data, metrics, and training protocols.
\subsection*{Order-defined one-dimensional benchmark}
\label{subsec:order_defined_benchmark}

We use a one-dimensional synthetic operator-learning benchmark with an explicitly controllable order parameter. The benchmark is used for order-controlled comparisons between cascaded KNCs and depth-stacked first-order baselines. On the periodic domain $s\in[0,1]$, the input-output mapping is defined in the Fourier domain as
\begin{equation}
    \hat{y}(k)=H_n(k)\hat{x}(k),
    \qquad
    H_n(k)=\frac{1}{\left(1+\tau^2 k^2\right)^n},
    \label{eq:1d_order_operator_fourier}
\end{equation}
where $x(s)$ is the input signal, $y(s)$ is the target response, $\tau>0$ is a scale parameter, and $n\in\mathbb{N}$ specifies the order of the operator family. Equivalently, the same mapping can be written in the spatial domain as
\begin{equation}
    \left(I-\tau^2\partial_{ss}\right)^n y=x.
    \label{eq:1d_order_operator_pde}
\end{equation}
In all experiments, we set $\tau=0.08$ and discretize each signal on a uniform grid with 256 points. The target orders are denoted as T1--T4, corresponding to first- through fourth-order instances of Eq.~\eqref{eq:1d_order_operator_pde}.

The operator family has an exact compositional form. From Eq.~\eqref{eq:1d_order_operator_fourier},
\begin{equation}
    H_n(k)=\bigl(H_1(k)\bigr)^n,
    \qquad
    H_1(k)=\frac{1}{1+\tau^2 k^2}.
    \label{eq:1d_factorization}
\end{equation}
Equivalently, if
\begin{equation}
    \mathcal{T}_n=
    \left(I-\tau^2\partial_{ss}\right)^{-n},
    \qquad
    \mathcal{T}_1=
    \left(I-\tau^2\partial_{ss}\right)^{-1},
\end{equation}
then
\begin{equation}
    \mathcal{T}_n=\mathcal{T}_1^{\,n}.
    \label{eq:1d_operator_composition}
\end{equation}
This factorization gives a controlled setting in which the target operator order can be varied without changing the spatial domain, grid resolution or data-generation procedure.

For comparison with the scan-wise KNC formulation, consider a single KNC with fixed coefficients,
\begin{equation}
    v_{t+1}=a v_t+b u_t,
    \qquad
    y_t=c v_{t+1}+d u_t,
    \label{eq:single_knc_frozen}
\end{equation}
where $t$ denotes the scan index. Its transfer function is
\begin{equation}
    H_{\mathrm{KNC}}^{(1)}(z)
    =
    d+\frac{cb}{1-a z^{-1}}.
    \label{eq:single_knc_transfer}
\end{equation}
An $N$-stage cascaded KNC gives
\begin{equation}
    H_{\mathrm{KNC}}^{(N)}(z)
    =
    \prod_{\ell=1}^{N}
    \left(
    d_\ell+\frac{c_\ell b_\ell}{1-a_\ell z^{-1}}
    \right).
    \label{eq:cascaded_knc_transfer}
\end{equation}
In the strictly recurrent case $d_\ell=0$, this reduces to
\begin{equation}
    H_{\mathrm{KNC}}^{(N)}(z)
    =
    \frac{
    \prod_{\ell=1}^{N} c_\ell b_\ell
    }
    {
    \prod_{\ell=1}^{N}(1-a_\ell z^{-1})
    }.
    \label{eq:cascaded_knc_strict}
\end{equation}
Thus, the benchmark provides an explicit operator-order axis for comparing a cascade formed inside one block with a network obtained by stacking first-order blocks.

The input signal $x(s)$ is generated as a mixture of low-frequency components, broadband oscillatory components and localized Gaussian pulses. This construction produces signals with both smooth global variation and localized high-variation structures. The target response $y(s)$ is generated analytically by applying the Fourier-domain transfer function in Eq.~\eqref{eq:1d_order_operator_fourier}. This avoids numerical PDE solver error and keeps the data-generation process deterministic.

We generate 12,000 training samples, 2,000 validation samples and 2,000 test samples. The three splits are generated using independent random seeds 42, 43 and 44, respectively. The CHONN variant used in this benchmark is implemented as an SS1DPassOnlyChain with expansion ratio 16 and state dimension $d_{\mathrm{state}}=8$. In the matched-order setting, the number of cascaded KNC stages is set to the target order $n$. The depth-stacked baseline is constructed by stacking first-order blocks, with each block containing only one KNC. For scaling comparisons, we additionally evaluate stacked high-order variants denoted by $2\times k$ and $3\times k$, where the first number denotes the intra-block order and the second denotes the number of stacked blocks.

All models are trained using AdamW with learning rate $1\times10^{-4}$ and batch size 32 for 80 epochs. A cosine annealing learning-rate schedule is used, and the training seed is fixed to 42. We evaluate physical-domain relative $L_2$ error, spectral-domain relative $L_2$ error~\cite{li2021fourier} and derivative relative $L_2$ error~\cite{cai2025efficient}:
\begin{equation}
    \mathrm{RelL2}_{\mathrm{point}}
    =
    \frac{
    \|y_{\mathrm{pred}}-y_{\mathrm{true}}\|_2
    }
    {
    \|y_{\mathrm{true}}\|_2
    },
    \label{eq:rel_l2_point_1d}
\end{equation}
\begin{equation}
    \mathrm{RelL2}_{\mathrm{spec}}
    =
    \frac{
    \|\mathcal{F}(y_{\mathrm{pred}})
    -
    \mathcal{F}(y_{\mathrm{true}})\|_2
    }
    {
    \|\mathcal{F}(y_{\mathrm{true}})\|_2
    },
    \label{eq:rel_l2_spec_1d}
\end{equation}
\begin{equation}
    \mathrm{RelL2}_{\mathrm{der}}
    =
    \frac{
    \|\nabla_s y_{\mathrm{pred}}-\nabla_s y_{\mathrm{true}}\|_2
    }
    {
    \|\nabla_s y_{\mathrm{true}}\|_2
    },
    \label{eq:rel_l2_der_1d}
\end{equation}
where $\mathcal{F}(\cdot)$ denotes the discrete Fourier transform and $\nabla_s$ denotes the finite-difference derivative along the one-dimensional coordinate.

\subsection*{Steady-state PDE operator-learning benchmarks}
\label{subsec:steady_pde_benchmarks}

We evaluate steady-state operator learning on two representative elliptic PDE benchmarks: Darcy flow and the Poisson equation. These experiments follow recent
operator-learning studies that learn PDE solution operators or non-local integral
operators directly from data~\cite{li2021fourier,goswami2022deep,zappala2024learning}. These tasks map an input field to a spatial solution field and therefore test whether the proposed cascaded Kirchhoff construction can improve globally coupled solution-operator approximation, rather than only short-range feature reconstruction.

\subsubsection*{Darcy flow benchmark}
\label{subsubsec:darcy_benchmark}

We first consider a two-dimensional Darcy flow benchmark, which represents a canonical steady-state elliptic problem in heterogeneous media. The governing equation is
\begin{equation}
-\nabla \cdot \big(a(x)\nabla u(x)\big)=f(x), 
\qquad x\in D,
\label{eq:darcy}
\end{equation}
subject to homogeneous Dirichlet boundary conditions
\begin{equation}
u(x)=0,\qquad x\in \partial D,
\label{eq:darcy_bc}
\end{equation}
where \(D=(0,1)^2\), \(a(x)\) denotes the spatially varying diffusion or permeability-related coefficient field, \(f(x)\) is the forcing term, and \(u(x)\) is the corresponding steady-state solution field. The learning objective is to approximate the coefficient-to-solution operator
\begin{equation}
\mathcal{G}^{\dagger}: a \mapsto u .
\label{eq:darcy_operator}
\end{equation}
Because the divergence-form elliptic operator couples local gradients, heterogeneous coefficients and boundary conditions, this benchmark provides a controlled test of non-local spatial coupling and boundary-sensitive solution recovery.

We use the Darcy-flow setting from PDEBench with forcing scale \(\beta=1.0\). In this setting, the forcing term is constant up to the prescribed scale parameter, while the coefficient field \(a(x)\) varies across samples. Rather than solving Eq.~\eqref{eq:darcy} directly as a static system, the reference solution is obtained by evolving the corresponding auxiliary parabolic equation
\begin{equation}
\partial_t u(x,t)-\nabla \cdot \big(a(x)\nabla u(x,t)\big)=f(x)
\label{eq:darcy_auxiliary}
\end{equation}
from a random-field initial condition until convergence to a steady state. Each sample is represented on a \(128\times128\) uniform grid. We use 800 samples for training, 100 samples for validation and 100 samples for testing.

CHONN is implemented by inserting second-order Cascaded Kirchhoff Blocks (CKBs) into a U-Net-style neural-operator backbone. The CKB modules are placed in the deepest encoder--decoder stages, where the receptive field is largest and the latent features encode macro-scale spatial structures. The CHONN model uses \(N_{\mathrm{layers}}=3\), \(N_{\mathrm{res}}=4\), \(N_{\mathrm{res\_neck}}=4\), channel multiplier 32, convolution kernel size 3 and filter size 6. This placement follows the architectural design in Fig.~\ref{fig4:physics}, allowing the network to combine hierarchical multi-scale feature extraction with explicit cascaded latent evolution.

For comparison, we evaluate FNO and U-Net baselines under the same dataset split and spatial resolution. The baseline batch size is set to 50. CHONN is trained using AdamW with learning rate \(1\times10^{-3}\), weight decay \(1\times10^{-3}\), and a StepLR schedule with step size 10 and decay factor \(\gamma=0.98\). We train for 1000 epochs with batch size 4 and random seed 0000. Validation performance is monitored on the held-out validation split, and the final model is evaluated on the test split.

The primary Darcy metric is the relative \(L_1\) error,
\begin{equation}
\mathrm{RelL1}
=
\frac{\|u_{\mathrm{pred}}-u_{\mathrm{true}}\|_1}
{\|u_{\mathrm{true}}\|_1}
\times 100\% ,
\label{eq:rell1}
\end{equation}
where \(u_{\mathrm{pred}}\) and \(u_{\mathrm{true}}\) denote the predicted and reference pressure fields, respectively. For the multi-metric analysis in Fig.~\ref{fig4:physics}, we additionally report RMSE, normalized RMSE, maximum pointwise error, boundary discrepancy and frequency-band errors. These complementary metrics evaluate global field accuracy, local error concentration, boundary-sensitive behavior and spectral consistency. The boundary discrepancy is computed on grid points adjacent to \(\partial D\), while frequency-band errors are computed after applying a two-dimensional discrete Fourier transform and grouping Fourier modes into low-, middle- and high-frequency bands. All reported error metrics are lower-is-better.

\subsubsection*{Poisson equation benchmark}
\label{subsubsec:poisson_benchmark}

We further evaluate the effect of CKB cascade order on a two-dimensional Poisson equation benchmark. The governing equation is
\begin{equation}
-\Delta u=f,\qquad \text{in } D,
\qquad u|_{\partial D}=0,
\label{eq:poisson}
\end{equation}
where \(D=[0,1]^2\), \(f(x,y)\) is the input source field, \(u(x,y)\) is the corresponding solution field, and \(\Delta=\partial_{xx}+\partial_{yy}\) is the two-dimensional Laplace operator. The learning objective is to approximate the solution operator
\begin{equation}
\mathcal{G}^{\dagger}: f\mapsto u .
\label{eq:poisson_operator}
\end{equation}
This benchmark is a steady-state function-to-function regression problem. Because the inverse Poisson operator is globally coupled, the task tests whether the model can recover long-range spatial dependencies and fine solution structures from a prescribed forcing field.

The source term is generated using a truncated sine-series expansion,
\begin{equation}
f(x,y)=\frac{\pi}{K^2}
\sum_{i,j=1}^{K}
a_{ij}(i^2+j^2)^{-r}
\sin(\pi i x)\sin(\pi j y),
\label{eq:poisson_source}
\end{equation}
where \(a_{ij}\sim \mathcal{U}[-1,1]\) are independently sampled coefficients and \(r=-0.5\). For in-distribution training and evaluation, we set \(K=16\). To evaluate out-of-distribution generalization to higher spectral complexity, we additionally construct a test set with \(K=20\). For each source field, the corresponding reference solution is obtained numerically, yielding paired samples \((f,u)\) for supervised operator learning.

All source and solution fields are discretized on a \(64\times64\) uniform grid over \(D=[0,1]^2\). The dataset contains 1024 training samples, 128 validation samples, 256 in-distribution test samples and 256 out-of-distribution test samples. Training fields are normalized to the range \([0,1]\), and the same normalization constants are applied to the validation and test sets.

For this benchmark, we use Poisson prediction as a cascade-order ablation to examine whether increasing the number of KNC stages improves solution-operator approximation. The original backbone without CKB is treated as the 0-order baseline. We then evaluate CKB variants from first order to fourth order by changing only the number of cascaded KNC stages while keeping the remaining backbone and training protocol unchanged. The CHONN variant uses \(N_{\mathrm{layers}}=4\), state dimension \(d_{\mathrm{state}}=16\) and width 16. For reference, the aligned CNO~\cite{raonic2023convolutional} baseline uses \(N_{\mathrm{layers}}=3\) and \(d_e=16\).

CHONN is trained using Adam with learning rate \(1\times10^{-3}\), weight decay \(1\times10^{-6}\), and a StepLR schedule with step size 10 and decay factor \(\gamma=0.98\). We train for 1000 epochs with batch size 16 and random seed 0000. Baseline models are evaluated using the corresponding reported setting with batch size 32 when applicable. The main cascade-order comparison keeps the training data, validation data, optimizer schedule and evaluation protocol fixed across all CKB orders.

The primary Poisson metric is the relative \(L_1\) error,
\begin{equation}
\mathrm{RelL1}
=
\frac{\|u_{\mathrm{pred}}-u_{\mathrm{true}}\|_1}
{\|u_{\mathrm{true}}\|_1}
\times 100\% .
\label{eq:poisson_rell1}
\end{equation}
We also report normalized RMSE and maximum mean-squared error to capture global solution accuracy and localized high-error regions:
\begin{equation}
\mathrm{nRMSE}
=
\frac{
\sqrt{\frac{1}{N}\sum_{p=1}^{N}
\left(u_{\mathrm{pred}}(p)-u_{\mathrm{true}}(p)\right)^2}
}
{
u_{\max}-u_{\min}
},
\label{eq:nrmse}
\end{equation}
and
\begin{equation}
\mathrm{MaxMSE}
=
\max_{p}
\left(u_{\mathrm{pred}}(p)-u_{\mathrm{true}}(p)\right)^2 ,
\label{eq:maxmse}
\end{equation}
where \(p\) indexes spatial grid points and \(N\) is the number of grid points. These metrics are used to evaluate both average solution-field fidelity and the model's ability to control local error peaks under increasingly high-order cascaded evolution.

\subsection*{Time-dependent PDE forecasting benchmarks}
\label{subsec:time_dependent_pde_benchmarks}

We evaluate time-dependent physical forecasting on two fluid-dynamics benchmarks: the two-dimensional shallow-water equations~\cite{takamoto2022pdebench} and the two-dimensional incompressible Navier--Stokes equations~\cite{li2021fourier}. In both settings, the model receives historical field observations and predicts future states.

\paragraph{Shallow-water equations.}
We use the two-dimensional shallow-water benchmark from PDEBench. The governing equations are
\begin{align}
    \partial_t h+\partial_x(hu)+\partial_y(hv)&=0,\\
    \partial_t(hu)+\partial_x\left(u^2h+\frac{1}{2}g_rh^2\right)
    +\partial_y(uvh)&=-g_rh\,\partial_x b,\\
    \partial_t(hv)+\partial_y\left(v^2h+\frac{1}{2}g_rh^2\right)
    +\partial_x(uvh)&=-g_rh\,\partial_y b,
\end{align}
where $h$ is the water depth, $u$ and $v$ are the horizontal velocity components, $b$ denotes the bathymetry, and $g_r$ is the gravitational acceleration.

The benchmark is instantiated as a radial dam-break problem on the square domain $\Omega=[-2.5,2.5]^2$. The initial water height is defined as
\begin{equation}
h(0,x,y)=
\begin{cases}
2.0, & \sqrt{x^2+y^2}<r,\\
1.0, & \sqrt{x^2+y^2}\ge r,
\end{cases}
\label{eq:swe_initial}
\end{equation}
where the radius $r$ is sampled from $\mathcal{U}(0.3,0.7)$. The simulations are generated using the PyClaw finite-volume solver. The dataset contains 1,000 samples, each discretized on a $128\times128$ grid with 100 temporal steps.

The model receives $T_{\mathrm{in}}=10$ historical frames and is trained with an unroll length of 20. The FNO baseline uses 12 Fourier modes and width 20. The CHONN variant inserts a cascaded evolution branch into the FNO-style backbone, with state dimension $d_{\mathrm{state}}=16$ and downsampling factor 2.

For FNO, U-Net and CHONN, we use AdamW with learning rate $1\times10^{-3}$ and StepLR with step size 100 and decay factor $\gamma=0.5$. These models are trained for 500 epochs with batch size 5 and random seed 0000.

\paragraph{Navier--Stokes rollout.}
We use the standard two-dimensional incompressible Navier--Stokes benchmark in vorticity form. The governing equation is
\begin{equation}
    \partial_t w(x,t)+u(x,t)\cdot\nabla w(x,t)
    =\nu\Delta w(x,t)+f(x),
    \qquad x\in(0,1)^2,\ t\in(0,T],
    \label{eq:ns_vorticity}
\end{equation}
subject to the incompressibility constraint
\begin{equation}
    \nabla\cdot u(x,t)=0,
    \qquad x\in(0,1)^2,\ t\in[0,T],
    \label{eq:ns_incompressibility}
\end{equation}
and the initial condition
\begin{equation}
    w(x,0)=w_0(x),
    \qquad x\in(0,1)^2.
    \label{eq:ns_initial}
\end{equation}
Here, $u$ denotes the velocity field, $w=\nabla\times u$ is the vorticity, $\nu$ is the viscosity coefficient and $f(x)$ is a fixed forcing term.

The initial vorticity field is sampled from a Gaussian random field,
\begin{equation}
    w_0\sim \mathcal{N}\left(0,\;7^{3/2}(-\Delta+49I)^{-2.5}\right),
    \label{eq:ns_grf}
\end{equation}
with periodic boundary conditions. The forcing is fixed as
\begin{equation}
    f(x)=0.1\left(\sin(2\pi(x_1+x_2))+\cos(2\pi(x_1+x_2))\right).
    \label{eq:ns_forcing}
\end{equation}
Reference solutions are generated using the stream-function formulation with a pseudo-spectral solver. A Poisson equation is solved in Fourier space to recover the velocity field; the vorticity is then differentiated, and the nonlinear transport term is evaluated in physical space and dealiased. Time integration is performed with a Crank--Nicolson update, where the nonlinear term is treated outside the implicit part. Simulations are generated on a $256\times256$ grid and downsampled to $64\times64$ for training and testing. The data-generation time step is $10^{-4}$, and the solution is recorded every $t=1$ time unit.

In our experiments, we use viscosity $\nu=1\times10^{-3}$ and spatial resolution $64\times64$. The standard setting uses $T_{\mathrm{in}}=10$ input frames. We additionally evaluate $3\rightarrow1$, $5\rightarrow1$ and $10\rightarrow1$ settings, where the model receives the previous $m$ vorticity frames and predicts the next frame. During evaluation, the one-step predictor is recursively applied to generate a 40-step rollout.

The dataset contains 1,000 training samples and 200 validation/test samples. The FNO baseline uses 12 Fourier modes and width 20. The CHONN variant follows the same FNO-style backbone but replaces the $w_0$ convolutional branch with the proposed cascaded state-space evolution operator using $d_{\mathrm{state}}=16$. Both FNO and CHONN are trained using Adam with learning rate $1\times10^{-3}$ and weight decay $1\times10^{-4}$. A StepLR schedule is used with step size 100 and decay factor $\gamma=0.5$. Models are trained for 500 epochs with batch size 20 and random seed 0000.

\subsection*{Large-scale visual recognition benchmark}
\label{subsec:imagenet_benchmark}

We evaluate large-scale visual recognition on the ImageNet-1K classification benchmark~\cite{deng2009imagenet}. ImageNet-1K contains approximately 1.28 million training images and 50,000 validation images from 1,000 object categories. Unlike the PDE benchmarks, this task does not provide an explicit governing equation, and is used to evaluate whether the proposed circuit-inspired high-order module can be integrated into general visual representation learning.

CHONN is implemented within a hierarchical visual state-space backbone by replacing the original first-order routing module with the proposed cascaded circuit-evolution module. The backbone follows a four-stage design with patch size 4, three input channels and 1,000 output classes. We evaluate Tiny, Small and Base model scales. The Tiny model uses embedding dimension 96 and stage depths $[2,2,6,2]$, giving stage widths $[96,192,384,768]$. The Small model uses embedding dimension 128 and stage depths $[2,2,4,2]$, giving stage widths $[128,256,512,1024]$. The Base model uses embedding dimension 128 and stage depths $[2,2,9,2]$, also giving stage widths $[128,256,512,1024]$.

All variants use a maximum drop-path rate of 0.2 and an MLP expansion ratio of 4.0. The patch-embedding stem consists of two convolutional layers with channel-wise normalization, producing an effective patch stride of 4. Spatial downsampling between adjacent stages is performed by strided convolution followed by channel-wise normalization. The classification head applies normalization, global average pooling and a linear classifier.

The CHONN branch keeps the hidden dimension of each visual block and uses a lightweight depthwise convolution before the selective Kirchhoff state update. Each Cascaded Kirchhoff Block contains two Kirchhoff Neural Cells by default. The cells are connected in series, so that the second cell receives the current block input together with the response produced by the previous cell. The two stage responses are then merged and used in the residual block output. This design preserves the serial high-order construction used in the PDE experiments while keeping the module compatible with a standard hierarchical vision backbone.

Images are trained at an input resolution of $224\times224$. Training uses bicubic interpolation, color jitter, RandAugment, random erasing, Mixup, CutMix and label smoothing. Because Mixup is enabled, models are trained with soft-target cross-entropy loss. During validation, images are resized to 256 and center-cropped to $224\times224$, followed by ImageNet mean--standard-deviation normalization. We report Top-1 and Top-5 accuracy on the ImageNet-1K validation set, with Top-1 accuracy used as the main metric.

Models are trained for 300 epochs using AdamW with weight decay 0.05, a cosine learning-rate schedule and 20 warm-up epochs. The base learning rate is linearly scaled with the effective global batch size. In the default setting, training uses distributed data parallelism on 8 GPUs with batch size 128 per GPU. Gradient clipping and automatic mixed precision are used, and an exponential moving average of model weights is maintained. For all models, we report Top-1 accuracy together with parameter count and FLOPs.

For the routing ablation, we compare four variants: original baseline, parallel expansion, serial cascade and serial cascade with circuit constraints. The original baseline denotes the standard visual state-space routing block without the intra-block cascade. The parallel expansion variant uses multiple evolution branches in parallel and merges their outputs, increasing branch capacity without introducing stage-to-stage propagation. The serial cascade variant connects multiple Kirchhoff Neural Cells sequentially within a block, allowing later cells to receive previous-stage responses. The serial cascade with circuit constraints further imposes the stability- and dissipation-oriented constraints used in the final CHONN implementation. For training-efficiency analysis, we report Top-1 accuracy at 30 epochs.

\section*{Data availability}

The datasets used in this study are publicly available. The Darcy flow dataset is the PDEBench \texttt{2D\_DarcyFlow\_beta1.0\_Train.hdf5} dataset and is available at \url{https://darus.uni-stuttgart.de/file.xhtml?fileId=133219&version=8.0}. The shallow-water equation dataset is the PDEBench radial dam-break dataset \texttt{2D\_rdb\_NA\_NA.h5} and is available at \url{https://darus.uni-stuttgart.de/file.xhtml?fileId=133021&version=8.0}. The Navier--Stokes dataset follows the standard Fourier Neural Operator benchmark setting and is available at \url{https://drive.google.com/drive/folders/1UnbQh2WWc6knEHbLn-ZaXrKUZhp7pjt-?usp=sharing}. The Poisson equation dataset is based on the CNO benchmark data \texttt{PoissonData\_64x64\_IN/OUT.h5} and is available at \url{https://zenodo.org/records/10406879}. ImageNet-1K is available from the official ImageNet repository under its license and access terms. The one-dimensional order-defined benchmark introduced in this work is generated analytically following the procedure described in the Methods section, and the corresponding data-generation scripts will be included in the code repository. Source data underlying the main figures will be provided with this manuscript.

\section*{Code availability}

Code will be made available to editors and reviewers during peer review and will be deposited in a public repository upon publication.

\bibliography{sn-bibliography}



\section*{Author contributions}

T.C., J.Z. and B.Z. conceptualized the study. T.C. and J.Y. contributed equally to this work. T.C. developed the methodology, performed the theoretical analysis, designed the experiments and analysed data. J.Y. implemented the models, performed the main experiments, curated data and evaluated results. J.Z. contributed to the methodology, design of the main experiments, and data analysis. L.Y. contributed to the ablation studies, manuscript organization and revision. J.L., D.D., C.X., L.H., T.W. and G.G. contributed to methodological validation, experimental analysis and manuscript revision. B.Z. guided the overall research direction and, together with L.Y. and J.Z., jointly supervised the work. All authors reviewed and approved the final manuscript.

\section*{Competing interests}

The authors declare no competing interests.

\end{document}